\pdfoutput=1

\documentclass[11pt]{article}

\usepackage{ACL2023}
\usepackage{times}
\usepackage{latexsym}
\usepackage[T1]{fontenc}

\usepackage[utf8]{inputenc}
\usepackage{times}
\usepackage{latexsym}
\usepackage{bm}
\usepackage{algorithm}
\usepackage{algorithmic}
\usepackage{multirow}
\usepackage{amsfonts}
\usepackage{color}
\usepackage{amsmath, mathtools}
\usepackage{microtype}

\usepackage{inconsolata}

\title{Large Language Models as Zero-Shot Keyphrase Extractors: \\A Preliminary Empirical Study}

\author{Mingyang Song, Xuelian Geng, Songfang Yao, Shilong Lu, Yi Feng, Liping Jing\thanks{\ \ Corresponding Author} \\
	Beijing Key Lab of Traffic Data Analysis and Mining \\
	Beijing Jiaotong University, Beijing, China \\
	{\tt mingyang.song@bjtu.edu.cn} \\
}

\begin{document}
\maketitle
\begin{abstract}
Zero-shot keyphrase extraction aims to build a keyphrase extractor without training by human-annotated data, which is challenging due to the limited human intervention involved. Challenging but worthwhile, zero-shot setting efficiently reduces the time and effort that data labeling takes. Recent efforts on pre-trained large language models (e.g., ChatGPT and ChatGLM) show promising performance on zero-shot settings, thus inspiring us to explore prompt-based methods. In this paper, we ask whether strong keyphrase extraction models can be constructed by directly prompting the large language model ChatGPT. Through experimental results, it is found that ChatGPT still has a lot of room for improvement in the keyphrase extraction task compared to existing state-of-the-art unsupervised and supervised models. We have released the related data\footnote{\url{https://github.com/MySong7NLPer/ChatGPT_as_Keyphrase_Extractor}}.
\end{abstract}

\section{Introduction}
Keyphrase extraction aims to extract a set of important phrases from unstructured text into structured data formats, which is a fundamental and crucial task in natural language processing \cite{2014survey,song_survey,song2023chatgpt}. Typically, keyphrase is beneficial for various applications due to its concise and precise expression \cite{li2023generate,pmlr,Tian_2023,hypersiamesenet,li2023unsupervised,HISUM,salewski2023zeroshot}. Working with an enormous amount of labeling data is always hectic, labor-intensive, and time-consuming. Hence, many organizations and companies rely on keyphrase extraction to automate manual work with zero / few-shot settings \cite{2021unsupervised, hguke, promptrank, hyperrank, agrank}.

Recent works \cite{agrawal2022,Wei2023ZeroShotIE} on large-scale pre-trained language models, such as GPT-3 \cite{llm_few}, InstructGPT \cite{ouyang2022training} and ChatGPT 2, suggest that large language models perform well in various natural language processing downstream tasks even without tuning the parameters. Therefore, it is a challenging question: Is it feasible to prompt large language models to achieve a zero-shot keyphrase extractor? Based on these clues, in this paper, we turn to ChatGPT and hypothesize that ChatGPT is born with the right abilities to deposit a zero-shot keyphrase extractor in an interactive mode.

In this paper, we mainly focus on verifying the performance of ChatGPT on four keyphrase extraction datasets and the ability of understanding long documents. Generally, long documents often necessitate complex processing strategies \cite{longke,ld1,ld2,ld3}. In the keyphrase extraction task, while it may be feasible to design improved algorithms for handling a multitude of candidate keyphrases in long documents, we believe that the effective incorporation of advanced features, especially those with background knowledge, can enhance the efficient discrimination between keyphrases and non-keyphrases, even when dealing with a vast number of candidate keyphrases. Simultaneously, the question of how to encode long documents is a topic worthy of exploration \cite{2014survey}. Consequently, we test the capacity of ChatGPT as a general large language model to handle long documents.  Through extensive experiments, it is found that ChatGPT still has a lot of room for improvement in the keyphrase extraction task compared to existing state-of-the-art unsupervised and supervised models.

\begin{table*}[t]
	\begin{center}
		\small
		\renewcommand\arraystretch{1.3}
		\renewcommand\tabcolsep{13pt}
		\begin{tabular}{l|c|c|c|c}
			\hline \hline 
			\textbf{\textsc{Test Set}} & \textbf{\textsc{Domain}}  & \textbf{\textsc{Type}} & \textbf{\textsc{\# Doc.}} & \textbf{\textsc{Avg. \# Words}} \\ \hline
			\textsc{Inspec} \cite{Inspec} &  Scientific Abstract & Short & 500 & 134.6 \\
			\textsc{DUC2001} \cite{Nus} &  News Article & Long & 308 & 847.2  \\
			\textsc{SemEval2010} \cite{SemEval} &  Scientific Article & Long & 100 & 1587.5  \\
			\hline
			\textsc{OpenKP} \cite{xiong19} &  Open Web Domain & Long & 6,616 & 900.4   \\
			\hline\hline
		\end{tabular}
	\end{center}
	\caption{\label{dataset} Statistics of used test sets. {\# Doc.} is the number of documents in the dataset.  {Avg. \# Words} is the average number of words for documents. {Present KPs (\%)} indicates the percentage of keyphrases, which are presented in the documents. {Note that this report uses all of the test data rather than sampling part from it}.}
\end{table*}
\begin{table}[t]
	\renewcommand\arraystretch{1.7}
	\begin{center}
		\scriptsize
		\begin{tabular}{p{0.5cm}p{6cm}}
			\hline\hline
			
			\multicolumn{2}{c}{\textsc{Prompts}} \\ \hline

			{\textbf{Tp}$1$}    & {\texttt{Extract keywords from this text:}} \texttt{[\textsc{Document}]} \\ \hline
			{\textbf{Tp}$2$}    & \texttt{Extract keyphrases from this text:} \texttt{[\textsc{Document}]} \\ 
			\hline\hline
			
		\end{tabular}
		\caption{\label{case} Two prompts are designed for chatting with ChatGPT to extract keyphrases from the text document.}
	\end{center}
\end{table}

\section{ChatGPT for Keyphrase Extraction}

\subsection{Evaluation Setting}
We briefly introduce the evaluation setting, which mainly includes the compared baselines, datasets, and evaluation metrics.
Note that each time a new query is made to ChatGPT, we clear conversations to avoid the influence of previous samples, which is similar to \citet{gangda}.

In this paper, we compare ChatGPT with several state-of-the-art unsupervised keyphrase extraction models: {TF-IDF} \cite{tfidf}, {YAKE} \cite{yake}, {TextRank} \cite{textrank}, {TopicRank} \cite{topicrank}, \textsc{PositionRank} \cite{positionrank}, {MultipartiteRank} \cite{multipartiterank}, {EmbedRank} \cite{embedrank}, {KeyGames} \cite{keygames}, {SIFRank} \cite{sifrank}, {AttentionRank} \cite{attentionrank}, {JointGL} \cite{2021unsupervised}, {MDERank} \cite{mderank}, {SetMatch} \cite{setmatch}, {HGUKE} \cite{hguke}, {AGRank} \cite{agrank}, {HyperRank} \cite{hyperrank}, {CentralityRank} \cite{centralityrank}, and {PromptRank} \cite{promptrank}. In addition to unsupervised models, we compare ChatGPT with two state-of-the-art supervised models: HyperMatch \cite{hypermatch} and KIEMP \cite{kiemp}. Furthermore, we also select the large language models ChatGLM-6B\footnote{\url{https://github.com/THUDM/ChatGLM-6B}} and ChatGLM2-6B\footnote{\url{https://github.com/thudm/chatglm2-6b}} for comparison. By default, the results in this report come from the ChatGPT version on 2023.03.01.

We evaluate ChatGPT and all the baselines on four datasets: \textsc{Inspec} \cite{Inspec}, \textsc{DUC2001} \cite{duc2001_singlerank}, \textsc{SemEval2010} \cite{SemEval}, and \textsc{OpenKP} \cite{xiong19}. Table~\ref{dataset} summarizes the statistics of used test sets.

Following previous studies \cite{kiemp, hypermatch, csl}, we adopt macro averaged F1@5 and F1@M to evaluate the quality of both present and absent keyphrases. When using F1@5, blank keyphrases are added to make the keyphrase number reach five if the prediction number is less than five. Similar to the previous studies \cite{icnlsp, hguke, setmatch, diversityrank}, we employ the Porter Stemmer to remove the identical stemmed keyphrases.

\subsection{Keyphrase Extraction Prompts}

ChatGPT requires human-designed prompts or instructions as guiding information to trigger its ability to achieve downstream tasks, i.e., extracting keyphrases. It should be noted that the style and quality of prompts can greatly affect the quality of ChatGPT keyphrase extraction. High-quality prompts often achieve better task performance than low-quality prompts. Figure~\ref{case} presents the prompts designed for extracting keyphrases, which is similar to \cite{song2023chatgpt}.

\subsection{Overall Performance}

As illustrated in Table~\ref{extraction}, it can be seen that ChatGPT achieves similar performance to the unsupervised method TF-IDF with simple prompts, and far surpasses the unsupervised method TextRank. Although there is a certain gap in the performance of ChatGPT compared to supervised methods HyperMatch and KIEMP, fine-tuning large models like ChatGPT through human-annotated data may make the performance far superior to existing supervised keyphrase extraction models. In addition, it is interesting that ChatGLM2 with only 6B parameters achieved comparable performance to ChatGPT.

\begin{table*}[h]
	\begin{center}
		\scriptsize
		\renewcommand\arraystretch{1.5}
		\renewcommand\tabcolsep{7pt}
		\begin{tabular}{l|ccc|ccc|ccc}
			\hline \hline 
			\multirow{2}{*}{\textbf{\textsc{Model}}} 
			& \multicolumn{3}{c|}{\textsc{\textbf{DUC2001}}} & \multicolumn{3}{c|}{\textsc{\textbf{Inspec}}} & \multicolumn{3}{c}{\textsc{\textbf{SemEval2010}}} \\
			& {F1@5} &  {F1@10} &  {F1@15} & {F1@5} &  {F1@10} &  {F1@15} & {F1@5} &  {F1@10} &  {F1@15} \\ \hline
			\multicolumn{10}{l}{\textsc{\textbf{Statistical Models}}}\\\hline
			TF-IDF \cite{tfidf} & 9.21 & 10.63 & 11.06 & 11.28 & 13.88 & 13.83 & 2.81 & 3.48 & 3.91 \\
			YAKE \cite{yake} & 12.27 & 14.37 & 14.76 & 18.08 & 19.62 & 20.11 & 11.76 & 14.4 & 15.19 \\
			\hline
			\multicolumn{10}{l}{\textsc{\textbf{Graph-based Models}}}\\\hline
			
			TextRank \cite{textrank} & 11.80 & 18.28 & 20.22 & 27.04 & 25.08 & 36.65 & 3.80 & 5.38 & 7.65 \\
			SingleRank \cite{duc2001_singlerank} & 20.43 & 25.59 & 25.70 & 27.79 & 34.46 & 36.05 & 5.90 & 9.02 & 10.58 \\
			TopicRank \cite{topicrank} & 21.56 & 23.12 & 20.87 & 25.38 & 28.46 & 29.49 & 12.12 & 12.90 & 13.54 \\
			PositionRank \cite{positionrank} & 23.35 & 28.57 & 28.60 & 28.12 & 32.87 & 33.32 & 9.84 & 13.34 & 14.33 \\
			MultipartiteRank \cite{multipartiterank} & 23.20 & 25.00 & 25.24 & 25.96 & 29.57 & 30.85 & 12.13 & 13.79 & 14.92 \\\hline
			
			\multicolumn{10}{l}{\textsc{\textbf{Embedding-based Models}}}\\\hline
			EmbedRankd2v \cite{embedrank} & 24.02 & 28.12 & 28.82 & 31.51 & 37.94 & 37.96 & 3.02 & 5.08 & 7.23 \\
			EmbedRanks2v \cite{embedrank} & 27.16 & 31.85 & 31.52 & 29.88 & 37.09 & 38.40 & 5.40 & 8.91 & 10.06 \\
			KeyGames \cite{keygames} & 24.42 & 28.28 & 29.77 & 32.12 & {40.48} & 40.94 & 11.93 & 14.35 & 14.62 \\
			SIFRank \cite{sifrank} & 24.27 & 27.43 & 27.86 & 29.11 & 38.80 & 39.59 & - & -& -\\
			SIFRank+ \cite{sifrank} & {30.88} & 33.37 & 32.24 & 28.49 & 36.77 & 38.82 & - & - & -\\
			AttentionRank \cite{attentionrank} & - & - & - & 24.45 & 32.15 & 34.49 & 11.39 & 15.12 & 16.66 \\
			JointGL \cite{2021unsupervised} & 28.62 & {35.52} & {36.29} & {32.61} & 40.17 & {41.09}& 13.02 & {19.35} & {21.72} \\
			MDERank \cite{mderank} & 23.31 & 26.65 & 26.42 & 27.85 & 34.36 & 36.40 & {13.05} & 18.27 & 20.35 \\
			SetMatch \cite{setmatch} & {31.19} & {36.34} & {38.72} & {33.54} & {40.63} & {42.11} & {14.44} & {20.79} & {24.18} \\
			HGUKE \cite{hguke} & {31.31} & {37.24} & {38.31} & {34.18}& \textbf{41.05} & \textbf{42.16} & {14.07} & {20.52}& {23.10}\\
			AGRank \cite{agrank} & - & - & - & \textbf{34.59} & 40.70 & 41.15 & 15.37 & 21.22 & 23.72 \\
			HyperRank \cite{hyperrank} & \textbf{32.68} & \textbf{39.18} & \textbf{40.21} & {33.35}& {40.79} & {42.12} & {14.79} & {21.33} & \textbf{24.20} \\
			CentralityRank \cite{centralityrank} & {31.63} & {37.77} & {38.77} & {32.99} & {40.93} & {41.73} & {15.51} & \textbf{21.39} & {23.83} \\
			PromptRank \cite{promptrank} & 27.39 & 31.59 & 31.01 & 31.73 & 37.88 & 38.17 & \textbf{17.24} & 20.66 & 21.35 \\
			\hline
			\multicolumn{10}{l}{\textsc{\textbf{LLM-based Models}}}\\\hline
			\multicolumn{1}{l|}{{ChatGPT (gpt-3.5-turbo) - \textbf{Tp}$1$}}
			& 19.29 & 23.32 & 22.98 & 22.95 & 30.57 & 33.87 & 13.25 & 15.94 & 17.12    \\
			\multicolumn{1}{l|}{{ChatGPT (gpt-3.5-turbo) - \textbf{Tp}$2$}}
			& 21.50 & 25.03 & 24.21 & 28.07 & 34.85 & 36.69 & 13.66 & 16.11 & 16.18    \\
			\multicolumn{1}{l|}{{ChatGLM-6B - \textbf{Tp}$2$}}
			& 11.57 & 11.72 & 11.09 & 14.52 & 18.85 & 20.17 & 6.69 & 8.71 & 9.14 \\
			\multicolumn{1}{l|}{{ChatGLM2-6B - \textbf{Tp}$2$}}
			& 15.08 & 15.44 & 13.96 & 25.07 & 30.09 & 31.56 & 13.12 & 13.79 & 13.94 \\
			\hline\hline
		\end{tabular}
	\end{center}
	\caption{\label{total} Performance on DUC2001, Inspec and SemEval2010 test sets. The best results are in bold.}
\end{table*}

\begin{table}[h]
	\scriptsize
	\centering
	\renewcommand\tabcolsep{7pt}
	\renewcommand\arraystretch{1.5}
	\begin{tabular}{c|ccc}
		\hline\hline
		\multirow{2}{*}{\textbf{\textsc{Model}}} & \multicolumn{3}{c}{\textsc{\textbf{OpenKP}}} \\ 
		& {F1@1}  & {F1@3} & F1@5 \\ \hline
		\multicolumn{4}{l}{\textsc{\textbf{Unsupervised Models}}} \\\hline
		\multicolumn{1}{l|}{{TF-IDF} \cite{tfidf}}
		& 19.6 & 22.3 & 19.6  \\ 
		\multicolumn{1}{l|}{{TextRank} \cite{textrank}}
		& 5.4 & 7.6 & 7.9  \\ \hline
		
		\multicolumn{4}{l}{\textsc{\textbf{PLM-based Models}}} \\\hline
		
		\multicolumn{1}{l|}{{HyperMatch} \cite{hypermatch}}
		& 36.4 & \textbf{39.4} & 33.8  \\ 
		\multicolumn{1}{l|}{{KIEMP} \cite{kiemp}}
		& \textbf{36.9} & 39.2 & \textbf{34.0}   \\
		\hline
		
		\multicolumn{4}{l}{\textsc{\textbf{LLM-based Models}}}\\\hline
		
		\multicolumn{1}{l|}{{ChatGPT (gpt-3.5-turbo) - \textbf{Tp}$1$}}
		& 16.5 & 21.1 & 17.4  \\
		\multicolumn{1}{l|}{{ChatGPT (gpt-3.5-turbo) - \textbf{Tp}$2$}}
		& 18.0 & 21.6 & 18.7  \\
		
		\multicolumn{1}{l|}{{ChatGLM-6B - \textbf{Tp}$2$}}
		& 11.5 & 8.5 & 7.1  \\
		\multicolumn{1}{l|}{{ChatGLM2-6B - \textbf{Tp}$2$}}
		& 16.0 & 11.0 & 8.6  \\
		\hline\hline
	\end{tabular}
	
	\caption{Results of keyphrase extraction on the OpenKP dataset. F1@3 is the main evaluation metric for this dataset \cite{xiong19}. The results of the baselines are reported in their corresponding papers. The best results are highlighted in bold.}
	\label{extraction}
\end{table}
\subsection{Long Document Understanding}

Generally, handling long documents poses a significant challenge in numerous natural language processing tasks, primarily due to the presence of intricate contexts within extended texts. This challenge is particularly pressing in keyphrase generation, where effectively comprehending long documents is paramount. Many of the existing keyphrase extraction baseline models, such as \cite{baseline}, \cite{kiemp}, and \cite{hypermatch}, necessitate the truncation of input documents to conform to the input constraints imposed by the underlying backbone, such as BERT \cite{bert}, resulting in a substantial loss of information. Simultaneously, extracting keyphrases that are contextually meaningful to the document is crucial in achieving semantic document understanding.

Consequently, we conduct experiments to measure the capability of ChatGPT to understand lengthy documents, as shown in Table~\ref{extraction} and Table~\ref{total}. From the results, it can be seen that compared with baselines, the extraction results of ChatGPT are not very ideal. Of course, various factors affect the performance of ChatGPT, and the results presented in this paper are only preliminary explorations. 
In addition, the extraction results of ChatGPT on the long document dataset SemEval2010 are significantly better than those on the Inspec and DUC2001 datasets, which verifies its excellent ability to understand long documents. As a next step, we plan to select longer documents, for example, documents containing approximately 4096 words, to serve as test cases.
\section{Conclusion}
In this paper, we conduct preliminary experiments and analysis on the keyphrase extraction task. Although ChatGPT has achieved good results in some natural language processing downstream tasks, there is still room for improvement in the keyphrase extraction task. Of course, due to limited available resources, this paper only conducted experimental verification under some basic settings, which may to some extent limit the performance of ChatGPT. In addition, there are many methods to optimize the extraction results of ChatGPT, such as designing more complex and high-quality prompts, constructing contextual samples as auxiliary information, and introducing supervised fine-tuning methods. 

In future research, utilizing large language models as external knowledge-assisted methods may be a hot topic in the keyphrase extraction task.

\bibliography{anthology}
\bibliographystyle{acl_natbib}
\end{document}